\documentclass[AMS,STIX1COL]{WileyNJD-v2}

\usepackage{verbatim, graphicx, mathtools}
\usepackage{amsmath}
\usepackage{array}
\usepackage{amsthm}
\usepackage{graphicx}
\usepackage{url}
\usepackage{xcolor} % for comments
\usepackage{bm} % for bold math symbols
\graphicspath{{./}}

\articletype{Article Type}%

\received{26 April 2016}
\revised{6 June 2016}
\accepted{6 June 2016}

\newcommand{\bx}{\bm{x}}

\newcommand{\ubar}[1]{\text{\b{$#1$}}}

\begin{document}

\title{Knot Selection in Sparse Gaussian Processes with a Variational Objective Function}

\author[1]{Nathaniel Garton*}

\author[2,3]{Jarad Niemi}

\author[3]{Alicia Carriquiry}

\authormark{NATHANIEL GARTON \textsc{et al}}

\address[1]{\orgdiv{Department of Statistics}, \orgname{Iowa State University}, \orgaddress{\state{Iowa}, \country{U.S.A.}}}

%\address[2]{\orgdiv{Org Division}, \orgname{Org Name}, \orgaddress{\state{State name}, \country{Country name}}}

%\address[3]{\orgdiv{Org Division}, \orgname{Org Name}, \orgaddress{\state{State name}, \country{Country name}}}

\corres{*Nathaniel Garton, \email{nmgarton@iastate.edu}}

%\presentaddress{This is sample for present address text this is sample for present address text}

\abstract[Summary]{Sparse, knot-based Gaussian processes have enjoyed considerable success as scalable approximations to full Gaussian processes. Certain sparse models can be derived through specific variational approximations to the true posterior, and knots can be selected to minimize the Kullback-Leibler divergence between the approximate and true posterior. While this has been a successful approach, simultaneous optimization of knots can be slow due to the number of parameters being optimized. Furthermore, there have been few proposed methods for selecting the number of knots, and no experimental results exist in the literature. We propose using a one-at-a-time knot selection algorithm based on Bayesian optimization to select the number and locations of knots. We showcase the competitive performance of this method relative to optimization of knots simultaneously on three benchmark data sets, but at a fraction of the computational cost.}

\keywords{sparse Gaussian processes, machine learning, knot selection, variational inference, nonparametric regression}

%\jnlcitation{\cname{%
%\author{Garton N.}, 
%\author{Niemi J.}, 
%\author{Carriquiry A.}} (\cyear{2020}), 
%\ctitle{Knot Selection in Sparse Gaussian Processes with a Variational Objective}, \cjournal{Stat Anal Data Min: The ASA Data Sci Journal}, \cvol{2020;00:1--6}.}

\maketitle

%\footnotetext{\textbf{Abbreviations:} ANA, anti-nuclear antibodies; APC, antigen-presenting cells; IRF, interferon regulatory factor}

\section{Introduction} \label{s:intro}
Gaussian processes (GPs) are a class of Bayesian nonparametric models with a plethora of uses such as
nonparametric regression and classification, spatial and time series modeling, density estimation, and 
numerical optimization and integration. Their use, however, is restricted to small data sets due to the 
need to store and invert an $N \times N$ covariance matrix, where $N$ is the number of observed data points. 
This leads to storage scaling $\mathcal{O}(N^2)$ and computation time scaling $\mathcal{O}(N^3)$. 

To address these computational challenges, there has been a large amount of literature on certain approximations to GPs, 
commonly called \textit{sparse} GPs, which achieve 
linear storage and time complexity in $N$ \citep{smola2001, williams2001, seeger2003, snelson2006, banerjee2008, finley2009, datta2016}. 
Many of these methods rely on a subset of input locations, which we refer to as knots, to induce marginal covariances between function values.
Models are defined so that the inverse of the approximating $N \times N$ covariance matrix, also called the precision matrix, is sparse. % CHANGE
That is, most of the elements of the precision matrix are zero, hence the justification for the name `sparse' GPs.   

Despite the success of these methods, one significant challenge in practice is selecting the number and locations of 
knots. One currently very popular practice is to optimize a predefined number of knots simultaneously alongside
covariance parameters with respect to some objective function using continuous optimization. The two most common objective 
functions are the marginal likelihood (or an approximation of it) \citep{snelson2006, guzman2008, cao2013, lobato2016}
and the evidence lower bound in the case that a variational inference approach is taken \citep{titsias2009, cao2013, hensman2015}. 
While this is often successful in practice, it requires the user to choose the number of knots, $K$, up front. One can opt to 
make $K$ as large as is computationally feasible, but this may not always be necessary to achieve accurate predictions;
 we will show this on some real data experiments. Further, as we will show, the computational burden associated with the 
continuous optimization may grow substantially due to a large number of additional parameters associated with the knots.

\cite{garton2020oat} proposed an efficient one-at-a-time (OAT) knot selection algorithm based on Bayesian optimization 
to select the number and locations of knots in sparse GPs when the objective function is the marginal likelihood. One aim
of their algorithm was to mitigate optimization issues often encountered when using the marginal likelihood as the 
objective function. However, they also found that even when the aforementioned optimization issues were not substantial,
 the OAT algorithm was able to effectively select knots so that 
the resulting models were competitively accurate as compared to doing simultaneous optimization. Furthermore,
the OAT algorithm tended to be several times faster than simultaneous optimization. 

In this paper, we extend the use of the novel OAT knot selection algorithm in \cite{garton2020oat} to 
the context of nonparametric regression and variational inference. We provide experimental results on 
three real data sets showing competitive accuracy of models selected using the OAT algorithm to those 
chosen via simultaneous optimization, but often at a lower computational cost. We also compare the performance
of the OAT algorithm when used with the evidence lower bound versus with the marginal likelihood as the objective function.

The remainder of this paper is as follows. In Section \ref{s:gps}, we briefly introduce Gaussian process regression. 
Section \ref{s:sparse_gps} introduces the class of knot-based, sparse GPs that we consider. Section \ref{s:vi} 
describes variational inference generally and in the context of the relevant sparse GP models. We also discuss
here some details regarding the evidence lower bound as the knot selection objective function, and we provide
an illustrative, one-dimensional regression example. In Section \ref{s:experiments}, we show experimental results
on three benchmark data sets, and in Section \ref{s:discussion} we conclude with a discussion.

\section{Gaussian Process Regression} \label{s:gps}
We assume that we have $N$ observations, $\{(y_i,x_i^{\top})\}_{i = 1}^{N}$, 
from a data set where each $y_i \in \mathbb{R}$ is the target of interest, 
and the values $x_i$ are vectors of input variables where 
$x_i \in \mathcal{X}$ and $\mathcal{X}$ is a compact subset of $\mathbb{R}^d$. 
%\jarad{How about $\mathcal{X}$ rather than $\mathcal{D}$?}
We suppose that over $\mathcal{X}$ there is an unobservable, 
real-valued function $f:\mathcal{X} \to \mathbb{R}$ 
taking values $f(x_i)$. 
We further suppose that the values of this function give the mean of the (conditional) 
distribution of the target random variable $Y_i$, and that the $Y_i$ random variables 
are conditionally independent given the $f(x_i)$. That is, we assume

\[ Y_i|f(x_i) \stackrel{ind}{\sim} \mathcal{N}(f(x_i), \tau^2), \]

\noindent where $\tau^2$ is variance due to random noise. Note that $\tau^2$ is also sometimes called a \textit{nugget}. 

We can use a GP as a prior distribution on the latent function. We denote this as $f(x) \sim \mathcal{GP}(m(x), k_{\theta}(x,x'))$,
where $m(x)$ is the mean function and $k_{\theta}(x,x')$ is the covariance function. We assume the covariance function
is parameterized by $\theta$. We will use $\bm{x} = \{ x_i \}_{i = 1}^{N}$ to denote the set of observed input locations,
 and we will use $\tilde{\bm{x}} = \{ \tilde{x}_i\}_{i = 1}^{J}$ to denote unobserved 
input locations at which we wish to predict the corresponding target values. The difference between $\bm{x}$ 
and $\tilde{\bm{x}}$ is that $f_{\tilde{\bm{x}}}$ depends on $Y$ only through $f_{\bm{x}}$. %CHANGE
A GP, by definition, is a collection of random variables such that any finite 
subcollection $f_{\bm{x}'} = (f(x'_1), ...,f(x'_M))^{\top} \sim \mathcal{N}_M(m_{\bm{x}'}, \Sigma_{\bm{x}'\bm{x}'})$ 
%\jarad{Using $n$ makes it seem like you need exactly the number of observations
%for this to be true. I suggest something other than $n$.}
where 
$m_{\bm{x}'} = (m(x'_1), ..., m(x'_M))^{\top}$ and the $ij$-th element of 
$\Sigma_{\bm{x}'\bm{x}'}(i,j) = k_{\theta}(x'_i, x'_j)$. In general, we will use notation $\Sigma_{\bm{x} \bm{x}'}$
to denote the matrix of covariances between elements of $f_{\bm{x}}$ and 
$f_{\bm{x}'}$ where $ij$-th element of $\Sigma_{\bm{x} \bm{x}'}(i,j) = k_{\theta}(x_i, x'_j)$.

Our assumed data model implies the following joint distribution for $(Y^\top, f_{\bm{x}}^\top)^\top$, 

\[ 
\begin{bmatrix} Y \\ f_{\bm{x}} \end{bmatrix} \sim \mathcal{N} \left( \begin{bmatrix} m_{\bm{x}} \\ m_{\bm{x}} \end{bmatrix}, 
\begin{bmatrix} 
\Sigma_{\bm{x} \bm{x}} + \tau^2 I & \Sigma_{\bm{x} \bm{x}} \\
\Sigma_{\bm{x} \bm{x}} & \Sigma_{\bm{x} \bm{x}}
\end{bmatrix} \right).
 \]

\noindent Similarly, we can write down the distribution for $(Y^\top, f_{\tilde{\bm{x}}}^\top)^\top$, which is
\[ 
\begin{bmatrix} Y \\ f_{\tilde{\bm{x}}} \end{bmatrix} \sim \mathcal{N} \left( \begin{bmatrix} m_{\bm{x}} \\ m_{\tilde{\bm{x}}} \end{bmatrix}, 
\begin{bmatrix} 
\Sigma_{\bm{x} \bm{x}} + \tau^2 I & \Sigma_{\bm{x} \tilde{\bm{x} }} \\
\Sigma_{\tilde{\bm{x}} \bm{x} } & \Sigma_{\tilde{\bm{x}} \tilde{\bm{x}} }
\end{bmatrix} \right).
 \]

\noindent Gaussian process prediction works by formulating the conditional distribution of $f_{\tilde{\bm{x}}}|Y$, 
which, using standard rules regarding multivariate Gaussian distributions, is the following 

\[ f_{\tilde{\bm{x}}}|Y \sim \mathcal{N}(m_{\tilde{\bm{x}}} + \Sigma_{\tilde{\bm{x}} \bm{x}} (\Sigma_{\bm{x} \bm{x}} + \tau^2 I )^{-1}(y - m_{\bm{x}}) \-\ , \-\  \Sigma_{\tilde{\bm{x}}\tilde{\bm{x}}} - \Sigma_{\tilde{\bm{x}} \bm{x}} (\Sigma_{\bm{x} \bm{x}} + \tau^2 I )^{-1}\Sigma_{\bm{x} \tilde{\bm{x} }}). \]

\section{Sparse, Knot-Based Gaussian Processes} \label{s:sparse_gps}
We discussed that GPs can be used as a prior distribution over functions. Importantly, however, 
GPs only directly impact inferences through a finite dimensional marginal distribution 
on relevant function values.
Sparse GPs  %CHANGE
are also used as prior distributions over the same relevant finite set of function values, but they have more appealing 
computational properties than full GPs \cite{candela2005}. Some, but not all, sparse GPs correspond to true functional priors 
\cite{candela2005}. Thus, sparse GPs 
are prior distributions which approximate the ideal, full GP prior. We explain this more precisely in 
the following paragraphs. It is worth noting that 
ordinarily the \textit{posterior} distribution of the latent function is of more interest than the prior. 
The variational inference method of \cite{titsias2009} that we discuss in Section \ref{s:vfe} directly 
specifies an approximation to the posterior of a full GP which corresponds to the approximate posterior 
resulting from one of the sparse priors discussed in this section. We will explain this in detail 
in Section \ref{s:vfe}.

The sparse Gaussian processes that we consider are all based on the assumption that conditional on a small subset of 
function values, the remaining function values in the \textit{training set} are independent. The input locations corresponding to this 
small set of function values have 
variously been referred to as knots \citep{banerjee2008, finley2009}, pseudo-inputs \citep{snelson2006}, 
or inducing points/inputs \citep{candela2005}. In the remainder, we will refer to them as knots. We will primarily examine only two 
sparse models called the deterministic training conditional (DTC) and the fully independent conditional (FIC) approximations, using  %CHANGE
naming conventions established by \cite{candela2005}. However, it will be useful to 
discuss an additional two models (deterministic inducing conditional (DIC) and fully 
independent training conditional (FITC)) to better understand this class of knot-based models 
\citep{candela2005}. We will explain the intuition behind these names in each of the relevant subsections. %CHANGE

Consider $K$ knots denoted by $\bm{x}^\dagger = \{x_k^{\dagger} \}_{k = 1}^{K}$. 
%\jarad{Why a set rather than a vector?}
%Let $p(f_{\bm{\tilde{x}}}, f_{\bm{x}}, f_{\bm{x}^\dagger})$ be the joint prior 
%for the latent function values at the prediction locations, observed input locations, as well as at the knots.
These are special input locations because they will induce the marginal covariances of all marginal 
function values. %CHANGE
 \cite{candela2005} showed that many of the approximate GP posteriors commonly used in practice
 \citep{smola2001, seeger2003, snelson2006, banerjee2008, finley2009} can be understood
as resulting from different kinds of approximate priors on $(f_{\tilde{\bm{x}}}, f_{\bm{x}}, f_{\bm{x}^\dagger})$. 
All approximate priors, $p(f_{\tilde{\bm{x}}}, f_{\bm{x}}, f_{\bm{x}^\dagger})$, are defined so that 

\[ p_{GP}(f_{\tilde{\bm{x}}}, f_{\bm{x}}, f_{\bm{x}^\dagger})  \approx p(f_{\tilde{\bm{x}}}, f_{\bm{x}}, f_{\bm{x}^\dagger}) = p(f_{\tilde{\bm{x}}}|f_{\bm{x}^\dagger})p(f_{\bm{x}} | f_{\bm{x}^\dagger})p_{GP}(f_{\bm{x}^\dagger}), \] %CHANGE

\noindent where we use the subscript $GP$ to specify the distribution implied by the full GP.
 All approximations require that $p(f_{\bm{x}}|f_{\bm{x}^\dagger}) = \Pi_{i = 1}^{N}p(f(x_i)|f_{\bm{x}^\dagger})$
where $f_{\bm{x}} = (f(x_1),\ldots,f(x_N))$. 
This results in a sparse precision matrix for $p(f_{\bm{x}}|f_{\bm{x}^\dagger})$ as well as for $p(f_{\bm{x}})$. 
%\jarad{I think you want the conditional distribution $p(f_{\bm{x}}|f_{\bm{x}^\dagger}).$}
%Further, all approximations set $p(f_{\bm{x}^\dagger}) = \mathcal{N}(m_{\bm{x}^\dagger}, \Sigma_{\bm{x}^\dagger \bm{x}^\dagger})$.
%That is, the distribution for the function at the knots is the same as that implied by the full GP.

The four approximations we discuss result from two possible decisions for distributions $p(f_{\bm{x}}|f_{\bm{x}^\dagger})$ and $p(f_{\tilde{\bm{x}}}|f_{\bm{x}^\dagger})$.
These approximations were all discussed in \cite{candela2005}. We will reproduce essentially the same exposition for clarity. 
These four models result from either correcting the covariance matrix of $f_{\bm{x}}|f_{\bm{x}^\dagger}$ to be the same as a full GP on the diagonal
or by using the full GP conditional distribution for $f_{\tilde{\bm{x}}}|f_{\bm{x}^\dagger}$. Table \ref{t:sparse_model_overview} 
shows the differences between the four sparse models we will consider in terms of whether or not the prior training and testing 
(co)variances match those of the full GP.
%\jarad{Provide an overview of the 4 models you will be discussing here.}

\begin{table}
\centering
\caption{Table showing whether or not certain marginal prior (co)variances implied by four sparse GP models 
match with the marginal prior (co)variances of the full GP.}
\begin{tabular}{|c|c|c|c|c|}
\hline
%& $V_{sparse}\left[ \phi_{x_i} \right] = V_{full}\left[ \phi_{x_i} \right]$ &
%$V_{sparse}\left[ \phi_{x^*} \right] = V_{full}\left[ \phi_{x^*} \right]$ \\
& Training covariances & Training variances & Test variances & Test covariances \\ 
\hline
DIC & NO & NO & NO & NO \\
DTC & NO & NO & YES & YES\\
FIC & NO & YES & YES & NO \\
FITC & NO & YES & YES & YES \\
\hline
\end{tabular}
\label{t:sparse_model_overview}
\end{table}

\subsection{Deterministic Inducing Conditional} \label{s:dic}

The first and simplest approximation has been called the subset of regressors \citep{rasmussen2006}, predictive process model \citep{banerjee2008}, 
and the deterministic inducing conditional (DIC) approximation \citep{candela2005}. We will
use the terminology of \cite{candela2005}. The DIC model assumes that the latent function is \textit{deterministic} once given %CHANGE
the function values at the knots. Any \textit{marginal} variance or covariance in the latent function is therefore \textit{induced} by the knots.
Let $\Sigma_{\bm{x} \bm{x}'}$ be the covariance matrix where the $ij$-th element is given by
$k_{\theta}(x_i,x_j')$ and define $\Psi_{\bm{x} \bm{x}'} \equiv \Sigma_{\bx \bx^{\dagger}} \Sigma_{\bm{x}^\dagger \bm{x}^\dagger}^{-1} \Sigma_{\bx^{\dagger} \bx'}$. 
Then the DIC approximation defines $p_{DIC}(f_{\bm{x}}|f_{\bm{x}^\dagger})$ and $p_{DIC}(f_{\tilde{\bm{x}}}|f_{\bm{x}^\dagger})$ as follows,

\begin{align*}
f_{\bm{x}} | f_{\bm{x}^\dagger} &\sim \mathcal{N}\left(m_{\bm{x}} + \Sigma_{x \bm{x}^\dagger} \Sigma_{\bm{x}^\dagger \bm{x}^\dagger}^{-1} (f_{\bm{x}^\dagger} - m_{\bm{x}^\dagger}), 0\right) \\
f_{\tilde{\bm{x}}} | f_{\bm{x}^\dagger} &\sim \mathcal{N}\left(m_{\tilde{\bm{x}}} + \Sigma_{ \tilde{\bm{x}} \bm{x}^\dagger} \Sigma_{\bm{x}^\dagger \bm{x}^\dagger}^{-1}(f_{\bm{x}^\dagger} - m_{\bm{x}^\dagger}), 0\right).
\end{align*} 

\noindent This, along with the marginal distribution $p(f_{\bm{x}^\dagger}) = \mathcal{N}(m_{\bm{x}^\dagger}, \Sigma_{\bm{x}^\dagger \bm{x}^\dagger})$ 
which will be consistent across all models, implies the following marginal distributions for $f_{\bm{x}}$ and $f_{\tilde{\bm{x}}}$
%\jarad{Why is the variance 0?}
%\jarad{You need a joint to imply a prior, not a conditional.}

\begin{align*}
p_{DIC}(f_{\bm{x}}) &= \mathcal{N}(m_{\bm{x}}, \Psi_{\bm{x} \bm{x}}) \\
p_{DIC}(f_{\tilde{\bm{x}}}) &= \mathcal{N}(m_{\tilde{\bm{x}}}, \Psi_{\tilde{\bm{x}} \tilde{\bm{x}} }).
\end{align*}

\cite{banerjee2008} showed that this approximation is an optimal approximation to the full GP in the sense that
for any location, $\tilde{x}$,
$E_{GP}\left[ \left. (f(\tilde{x}) - g(f_{\bx^\dagger}))^2 \right| f_{\bm{x}^\dagger} \right]$ is minimized when

\[ g(f_{\bm{x}^\dagger}) = m_{\tilde{x}} + \Sigma_{ \tilde{{x}} \bm{x}^\dagger} \Sigma_{\bm{x}^\dagger \bm{x}^\dagger}^{-1}(f_{\bm{x}^\dagger} - m_{\bm{x}^\dagger}). \]
%\jarad{It's not clear what $f(\tilde{\bm{x}})$ is. I'm guessing $\tilde{\bm{x}}$ is a single predictive location.} 

\noindent The expectation here is taken with respect to the full GP. Despite this optimal property, using this approximation
tends to result in the underestimation of posterior function variances. This is because the prior GP variances 
for the DIC model are smaller than for the full GP. To see this, note that 
for the full GP, $V_{GP}\left[ f_{\bm{x}} | f_{\bm{x}^\dagger} \right] = \Sigma_{\bm{x} \bm{x}} - \Psi_{\bm{x} \bm{x}}$.
 However, note that $V_{DIC}\left[ f_{\bm{x}} | f_{\bm{x}^\dagger} \right] = \Psi_{\bm{x} \bm{x}}$. 
%\jarad{This seems like the correct variance.}
Conditional variances are nonnegative implying that the diagonal elements of $\Psi_{\bm{x} \bm{x}}$ are smaller than the corresponding elements of $\Sigma_{\bm{x} \bm{x}}$ \citep{banerjee2008}. 
%\jarad{So?}
The same is true of predictive variances at unobserved locations $\tilde{\bm{x}}$.

\subsection{Deterministic Training Conditional} \label{s:dtc}

The variance underestimation problem has led to two modifications to the DIC model. The first was discussed in \cite{seeger2003}, which 
involved a different distribution for $p(f_{\tilde{\bm{x}}}|f_{\bm{x}^\dagger})$ resulting
in a model they call projected latent variables. \cite{candela2005} refer to this model as the deterministic training conditional (DTC) approximation. Whereas the DIC model assumed all function values were deterministic given the function values at the knots, the DTC model assumes that this is only true of function values at \textit{training} data input locations $x$. However, the function values at $\tilde{\bm{x}}$ are not assumed to be deterministic conditional on the function values at the knots. 
Specifically, this approximation assumes that

\[ f_{\tilde{\bm{x}}}|f_{\bm{x}^\dagger} \sim \mathcal{N}(m_{\tilde{\bm{x}}} + \Sigma_{ \tilde{\bm{x}} \bm{x}^\dagger} \Sigma_{\bm{x}^\dagger \bm{x}^\dagger}^{-1}(f_{\bm{x}^\dagger} - m_{\bm{x}^\dagger}) , \Sigma_{\tilde{\bm{x}} \tilde{\bm{x}} } - \Psi_{\tilde{\bm{x}} \tilde{\bm{x}} }). \]

\noindent This is the exact distribution for $f_{\tilde{\bm{x}}}|f_{\bm{x}^\dagger}$ if one were to use the full GP. Thus, $p_{DIC}(f_{\bm{x}}|f_{\bm{x}^\dagger}) = p_{DTC}(f_{\bm{x}}|f_{\bm{x}^\dagger})$, but $p_{DIC}(f_{\tilde{\bm{x}}}|f_{\bm{x}^\dagger}) \neq p_{DTC}(f_{\tilde{\bm{x}}}|f_{\bm{x}^\dagger}) = p_{GP}(f_{\tilde{\bm{x}}}|f_{\bm{x}^\dagger})$.

%\jarad{Consider simply introducing the 4 models with the relevant distributions,
%but waiting to compare the models until later. Perhaps it would actually be better
%to introduce a table quite early that consists of the distributions you are 
%interested in comparing.}

\subsection{Fully Independent Conditional} \label{s:fic}
The second modification to the DIC model was suggested independently in both \cite{snelson2006} and \cite{finley2009}
and was called a sparse pseudo-input GP and a modified/bias-corrected predictive process model in the two sources, respectively.
\cite{candela2005} refer to this model as the fully independent conditional (FIC) approximation. By contrast to the DIC approximation, 
the FIC model does not assume that function values are deterministic conditional on the function values at the knots, but it does assume that
function values are \textit{conditionally independent} and have conditional variances matching that of the full GP.   

This approximation makes modifications to both $p_{DIC}(f_{\bm{x}}|f_{\bm{x}^\dagger})$ and $p_{DIC}(f_{\tilde{\bm{x}}}|f_{\bm{x}^\dagger})$ as compared to 
the distributions considered by the DIC model. FIC assumes the following conditional distributions for $f_{\bm{x}}$ and $f_{\tilde{\bm{x}}}$,

\begin{align*}
f_{\bm{x}}|f_{\bm{x}^\dagger} &\sim \mathcal{N}\left(m_{\bm{x}} + \Sigma_{ \bm{x} \bm{x}^\dagger} \Sigma_{\bm{x}^\dagger \bm{x}^\dagger}^{-1}(f_{\bm{x}^\dagger} - m_{\bm{x}^\dagger}) , \text{diag}(\Sigma_{ \bm{x} \bm{x}} - \Psi_{\bm{x} \bm{x}})\right) \\
f_{\tilde{\bm{x}}}|f_{\bm{x}^\dagger} &\sim \mathcal{N}\left(m_{\tilde{\bm{x}}} + \Sigma_{ \tilde{\bm{x}} \bm{x}^\dagger} \Sigma_{\bm{x}^\dagger \bm{x}^\dagger}^{-1}(f_{\bm{x}^\dagger} - m_{\bm{x}^\dagger}) , \text{diag}(\Sigma_{\tilde{\bm{x}} \tilde{\bm{x}} } - \Psi_{\tilde{\bm{x}} \tilde{\bm{x}} })\right).
\end{align*}

\noindent This implies the following marginal distributions for $f_{\bm{x}}$ and $f_{\tilde{\bm{x}}}$,

\begin{align*}
p_{FIC}(f_{\bm{x}}) &= \mathcal{N}(m_{\bx} , \text{diag}(\Sigma_{\bx \bm{x}} - \Psi_{\bm{x} \bm{x}}) + \Psi_{\bm{x} \bm{x}}) \\
p_{FIC}(f_{\tilde{\bm{x}}}) &= \mathcal{N}(m_{\tilde{\bm{x}}} , \text{diag}(\Sigma_{\tilde{\bm{x}} \tilde{\bm{x}} } - \Psi_{\tilde{\bm{x}} \tilde{\bm{x}} }) + \Psi_{\tilde{\bm{x}} \tilde{\bm{x}} }).
\end{align*}

\noindent Thus, the FIC model assumes the same prior variances as the full GP, but the prior covariances are now different.

\subsection{Fully Independent Training Conditional} \label{s:fitc}
The final approximation we mention was first explicitly discussed in \cite{candela2005} and named the fully independent training conditional (FITC) model.
This approximation modifies the FIC model so that the predictive covariances match that of the full GP. That is, $f_{\tilde{\bm{x}}}|f_{\bm{x}^\dagger}$ is assumed
to have the following distribution 

\[ f_{\tilde{\bm{x}}}|f_{\bm{x}^\dagger} \sim \mathcal{N}(m_{\tilde{\bm{x}}} + \Sigma_{ \tilde{\bm{x}} \bm{x}^\dagger} \Sigma_{\bm{x}^\dagger \bm{x}^\dagger}^{-1}(f_{\bm{x}^\dagger} - m_{\bm{x}^\dagger}) , \Sigma_{\tilde{\bm{x}} \tilde{\bm{x}} } - \Psi_{\tilde{\bm{x}} \tilde{\bm{x}} }). \]

\noindent Thus, we have that $p_{FIC}(f_{\bm{x}}|f_{\bm{x}^\dagger}) = p_{FITC}(f_{\bm{x}}|f_{\bm{x}^\dagger})$, but $p_{FIC}(f_{\tilde{\bm{x}}}|f_{\bm{x}^\dagger}) \neq p_{FITC}(f_{\tilde{\bm{x}}}|f_{\bm{x}^\dagger}) = p_{GP}(f_{\tilde{\bm{x}}}|f_{\bm{x}^\dagger})$. 
%\jarad{I guess this notation works as you are equating functions.}

In the remainder, we will focus on the DTC and the FIC approximations. This is because we will see that the posterior distribution for $f_{\tilde{\bm{x}}}$ resulting from the DTC 
prior can be derived as the marginal of an optimal posterior approximation to $p_{GP}(f_{\tilde{\bm{x}}} , f_{\bm{x}} , f_{\bm{x}^\dagger}|y)$ in a sense that we will discuss in Section \ref{s:vfe}. Also, we are primarily interested in marginal predictive distributions, which are the same for the FIC and FITC models. 

%\jarad{I'm confused why all 4 were introduced only two models disregarded at this point. Perhaps it is just wording. Instead of saying ``we will only discuss'' perhaps say something like ``we will focus discussion on DTC and FIC acknowledging that marginal predictive distribution for FIC and FITC are equivalent, so that any statement made about FIC is also true of FITC.''}

\section{Variational Inference} \label{s:vi}
 In this section, we discuss variational inference (VI) in a general context, and in Section \ref{s:vfe}
we discuss the particular approximation relevant for GP regression.
Variational inference is an analytical, optimization-based method for approximating probability
distributions \citep{blei2017}. The goal of VI is to approximate a potentially intractable distribution 
$P$ defined on $\mathcal{Z}$ with a \emph{variational distribution}, $Q$. It is standard to assume that 
$P$ and $Q$ have probability densities $p$ and $q$, respectively, with respect to some probability 
measure $\mu$. We then define our objective function to be

\[ \mathcal{D}(Q||P) = \int_{\mathcal{Z}}{ q(z)\log \frac{q(z)}{p(z)} d\mu(z)}, \]

\noindent the Kullback-Leibler (KL) divergence of $P$ with respect to $Q$. We will consider this objective function in the context
of trying to approximate posterior distributions of some parameters $Z$ given observed data, $Y$. 
%\jarad{Is $X$ being used because the ``parameters'' are the knots? Are there additional parameters?}
Going forward, we will write $p(z|y)$ instead of $p(z)$ to make this explicit.
%\jarad{Why not do this right away? At this point it isn't clear exactly what $Y$ is.}

The KL divergence above is often not analytically tractable. 
\cite{jordan1999}, however, showed that minimizing the above KL
divergence is equivalent to maximizing a lower bound on the log-likelihood, commonly called the \textit{evidence lower bound} (or ELBO). 
%\jarad{Equivalent? Or the ELBO provides a lower bound?}
We reproduce this derivation as it is shown in
\citep{blei2017}. The KL divergence can be written as
\begin{align*}
 \mathcal{D}(Q||P) &=  E\left[ \log q(z) \right] -  E\left[ \log p(z,y) \right] + E\left[ \log p(y) \right] \\
&= E\left[ \log q(z) \right] -  E\left[ \log p(z,y) \right] + \log p(y),
 \end{align*}

\noindent where expectations are with respect to the distribution $Q$. 
By rearranging terms,  we see that 
\begin{align*}
\log p(y) &=  \mathcal{D}(Q||P) +  E\left[ \log p(z,y)  - E\left[ \log q(z) \right] \right] \\
 &\geq E\left[ \log p(z,y) \right] - E\left[ \log q(z) \right] \\
&= ELBO(q). 
\end{align*}

\noindent Thus, we see that by maximizing $ELBO(q)$ with respect to the distribution $q$, 
we minimize $\mathcal{D}(Q||P)$ since $\log p(y)$ is not a function of $q$.
For example, when $\log p(y) =  E\left[ \log p(x,y) \right] - E\left[ \log q(x) \right]$, it must be that $ \mathcal{D}(Q||P) = 0$ which implies that $P = Q$. 
In general, any arbitrary $Q$ need not result in an analytically tractable expression for the ELBO. However,
typically $q(z)$ and $p(z,y)$ will have analytical expressions, but the expectations may be challenging or impossible to compute analytically.

\subsection{Variational Inference in Sparse GPs} \label{s:vfe}
\cite{titsias2009} showed how the approximate posterior, $p_{DTC}(f_{\tilde{\bm{x}}}|y)$, can be derived by using a predictive distribution
that can be written as $\int p_{GP}(f_{\tilde{\bm{x}}}|f_{\bm{x}^\dagger})h^*(f_{\bm{x}^\dagger}) df_{\bm{x}^\dagger}$, 
where $h^*(f_{\bm{x}^\dagger}) = p_{DTC}(f_{\bm{x}^\dagger}|y)$ is
 the marginal distribution resulting from the optimal variational approximation to $p_{GP}(f_{\bm{x}}, f_{\bm{x}^\dagger}|y)$
 in the class of distributions, $\mathcal{Q}$, with densities $q$ that can be written as 
\[q(f_{\bm{x}}, f_{\bm{x}^\dagger}) = p_{GP}(f_{\bm{x}}|f_{\bm{x}^\dagger})h(f_{\bm{x}^\dagger}). \] 

\noindent Here, note that $h$ is considered to be a ``free form" variational distribution for $f_{\bm{x}^\dagger}$, meaning that it is not restricted to be from any specific distributional family. 
\cite{seeger2003} derives essentially the same result while pursuing the goal of finding and 
justifying a sparse likelihood approximation. 
We reproduce essentially the same derivation of the optimal variational 
distribution and the corresponding ELBO in Appendix \ref{sec:optimal}. 
The ELBO arising from this optimal variational approximation is given by

\[ ELBO(q^*) = \log \left[ \mathcal{N}(y \-\ ; \-\ m_{\bm{x}} , \Psi_{\bm{x} \bm{x}} + \tau^2I )  - \frac{1}{2\tau^2} Tr\left( V_{GP}\left[ f_{\bm{x}}|f_{\bm{x}^\dagger} \right] \right)  \right], \]

\noindent where we use $q^*$ to denote the optimal variational distribution.

Using the optimal variational approximation and ELBO, 
derivatives of the ELBO are taken with respect to covariance parameters and the knots.
These derivatives can be used to optimize the ELBO using a gradient-based optimization routine. 
In keeping with terminology in \cite{bauer2016}, we will refer to the model resulting from this variational approximation in combination with using the ELBO for model selection the \textit{variational free energy} (VFE) model.

\subsection{Knot Selection Using the ELBO} \label{s:knot_selection}
The ELBO is an appealing objective function for knot selection because it
never decreases with an addition of a new knot \citep{titsias2009, bauer2016}. 
To gain some intuition for this, first recall that maximizing the ELBO is equivalent to 
minimizing the KL divergence between the approximate and the full posterior. At a high level, 
adding knots results in a prior covariance matrix in the sparse model that better approximates 
the prior covariance matrix in the full GP model, and so the KL divergence between the two 
posteriors will be smaller. More concretely, note that the ELBO is the sum of two terms: 
the first is the marginal likelihood of the DTC/DIC model, and the second is a strictly negative 
term consisting of the negative (scaled) sum of the conditional variances of $f_{\bm{x}}$ given $f_{\bm{x}^{\dagger}}$ 
according to the full GP. The first term measures how well the model fits the data, but it doesn't depend at all 
on the full GP posterior that we are trying to approximate. The second term does not depend on the data, but 
it does depend on the full GP posterior (through the full GP prior). Thus, it is the second term that must encourage 
the approximate posterior to resemble that of the full GP. Indeed, $V_{GP}\left[ f_{\bm{x}} | f_{\bm{x}^\dagger} \right]$ 
can only decrease or remain constant as the number of knots grows. The fact that the change in the 
second term in the ELBO offsets any decrease in the first term is nontrivial, and we refer curious 
readers to \cite{bauer2016} for the proof.

\begin{figure}[htbp!]
\centering
 \includegraphics[width = 0.6\textwidth, height = 0.275\textheight]{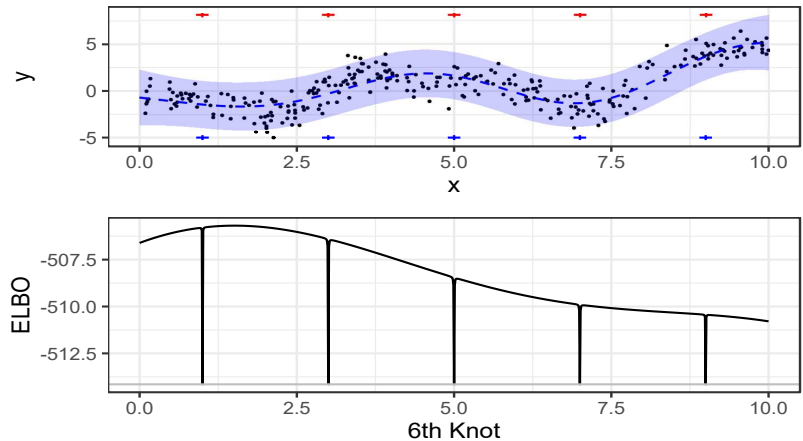}
 \caption{The top panel shows the fit from a five knot VFE model, 
while the bottom panel shows the ELBO values as a function of the location of a single, sixth knot with first five
 knots (blue and red $+$) held fixed. The ELBO value for the model without the sixth knot 
is plotted as a horizontal dashed line.
 }
 \label{fig:vi_spike_plot}
\end{figure}

%The second reason is that placing knots at each observed input location recovers the full GP log-likelihood \citep{titsias2009}. 
Unfortunately, adding knots one-at-a-time can be tricky in practice. 
An intuitively reasonable method for selecting knots and covariance parameters %CHANGE
might be to first initialize some small set of knots and covariance parameter values. 
One could then consider adding a knot followed by continuous optimization of the ELBO 
with respect to either the covariance parameters exclusively or the covariance parameters as well as the added knot. 
However, Figure \ref{fig:vi_spike_plot} shows a phenomenon discussed in 
\cite{bauer2016} where spikes in the ELBO exist whenever a new knot is placed 
directly on top of a previously existing knot.
Further, \cite{bauer2016} also note that the addition of a small noise variance of $f(x)$,
often necessary for numerical stability of matrix inverses,
results in a widening of these spikes. This causes suboptimal local
maxima, which can be sufficient to disrupt an optimization algorithm.

\cite{titsias2009} suggested the possibility of greedily adding a knot by choosing the value
that maximized improvement to the ELBO over some small random sample of observed data locations. 
While this may often work reasonably well in practice, there may be more efficient ways of searching 
the observed data locations. 
\cite{garton2020oat} proposed using Bayesian optimization
to efficiently propose a new knot which is then optimized alongside covariance 
parameters holding previous knots fixed using gradient based methods. %CHANGE 
\cite{garton2020oat} showed that compared to optimization of
all knots simultaneously, their OAT knot selection algorithm was often at least as accurate but was usually
many times faster. 
Thus, we propose using a slightly modified version of the OAT method to select knots using the ELBO
from the VFE method as the objective function. Note that this requires a covariance function 
that is differentiable in the knot locations. %CHANGE
The only difference between our implementation here 
and the implementation in \cite{garton2020oat} is that we do not 
condition on the values of the ELBO 
when the new knot is located in the same spot as an existing knot in the 
Bayesian optimization knot proposal function. 
As in \cite{garton2020oat}, 
we refer to the OAT algorithm that uses Bayesian optimization for the proposal
function as the OAT-BO algorithm. 
Because we are primarily concerned with regression problems, %CHANGE
in which the true latent function can reasonably be assumed to be fairly smooth, 
we consider using covariance functions resulting in smooth GP realizations. 
Furthermore, our knot selection algorithm requires that the covariance function 
is at least once differentiable in the knot locations. Thus, in every application 
we use the squared exponential covariance function, 
$k_\theta(x,x') = \sigma^2 e^{\frac{-||x - x'||^2}{2\ell^2}}$. However, 
one could certainly consider using any other covariance function that is once 
differentiable in the knot locations. 

%We wish to 
%avoid ever simultaneously optimizing all knots at once, especially when $d$ is moderate to large,
% because gradient evaluations scale $K\mathcal{O}(dNK^2)$,
%whereas optimizing only one knot at a time costs $\mathcal{O}(dNK^2)$.  

\begin{figure}[htbp!]
\centering
 \includegraphics[width = 0.8\textwidth, height = 0.3\textheight]{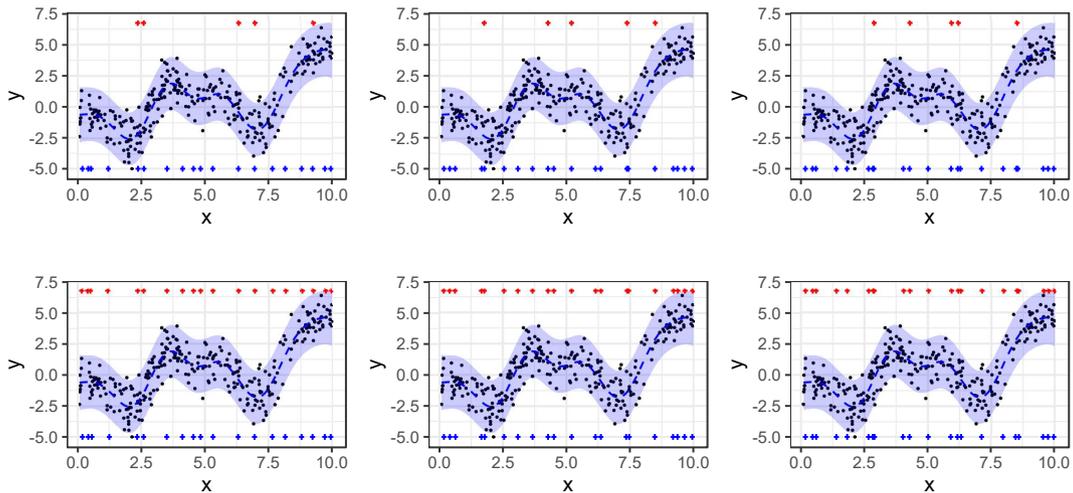}
 \caption{VFE model fits to a 300 observation synthetic, one dimensional regression 
using the OAT-BO algorithm (top row) and refinements to the placed knots through simultaneous
optimization (bottom row). Initial knots  (red $+$) and final knots (blue $+$) are shown on the top and bottom 
of each plot, respectively. 
%\jarad{Any chance you can facet?}
 }
 \label{fig:gaussian_example}
\end{figure}

As an illustrative example, Figure \ref{fig:gaussian_example} shows results on a synthetic, one dimensional regression problem with 300 observations.
% The top row of plots shows fits of the VFE model using the OAT-BO algorithm for three different random knot initializations.
% The bottom row shows fits of the VFE model after refining knots selected by the OAT-BO algorithm by simultaneously 
% optimizing knots and covariance parameters.  
We see that the OAT-BO algorithm selects knots roughly uniformly across 
the x-axis and selects roughly the same numbers of knots. 
We also see that the refinements to the knots placed by the OAT-BO
algorithm in the bottom row are minimal. 
Thus, in this case, the OAT-BO algorithm appears to have placed knots 
near a local maximum. 
The predictions and uncertainties from each fit looks nearly identical.

\section{Experiments} \label{s:experiments}
In this section, we compare the OAT-BO algorithm to several alternatives for knot selection on three publicly available data sets. 
In all experiments, 
we test the OAT-BO algorithm in a VFE model, 
the OAT-BO algorithm in an FIC model where the 
model selection objective function is the marginal likelihood, 
the OAT algorithm using the best-of-random-subset (abbreviated as `RS') proposal as in \cite{garton2020oat} in a VFE model, 
and a refinement of the fit of the VFE model selected through the OAT-BO algorithm by
simultaneously optimizing all knots and covariance parameters. 
In every model, we add a small nugget to the latent function
to ensure that the relevant inverses are numerically stable. 
Knots for all models, except for the VFE refinement, 
were initialized using k-means clustering. 
Covariance parameters in all models were initialized to the same values. 
The maximum number of knots allowed by all OAT algorithms was set to 80. 
Further, the number of knots in the simultaneously 
optimized models were set to be equal to the number found by the OAT-BO algorithm.
Lastly, all gradient based optimizations were done using ADADELTA \cite{zeiler2012}, 
as in \cite{garton2020oat}. \texttt{R} \cite{R} code to reproduce all results in this work is available 
as a package called \texttt{sparseRGPs} available at \url{https://github.com/nategarton13/sparseRGPs}.

We use the same, slightly modified versions of canonical performance metrics in 
\cite{garton2020oat},
reflecting the fact that we are only interested in marginal predictive densities. The
two main metrics we consider are common to all of our experiments. The first metric
is the median negative log-probability (MNLP), which is calculated as 
\[ 
MNLP = \text{median}_{i \in 1, ..., N_{test}} \{ -\log p(\tilde{y}_i|\bm{x}^{\dagger}, \hat{\theta}, y) \}.
\]
Lower MNLP values correspond to more accurate marginal predictive densities. 
The second metric we calculate is standardized root mean squared error (SRMSE), which is calculated by averaging the squared
differences between predictions and the test data and normalized by the sample standard deviation on the test set. That is,
\[ SRMSE = \sigma_{\tilde{y}}^{-1} \sqrt{ \frac{1}{N_{test}} \sum_{i = 1}^{N_{test}} (E\left[f(\tilde{\bm{x}}_i)|Y\right] - \tilde{y}_i )^2}, \]
where $\sigma_{\tilde{y}}^2 = \frac{1}{N_{test} - 1}\sum_{i = 1}^{N_{test}}(\tilde{y}_i - \ubar{\tilde{y}})^2$, $\ubar{\tilde{y}} = \frac{1}{N_{test}} \sum_{i = 1}^{N_{test}}\tilde{y}_i$, and $\tilde{y}$ is the vector of test set target values.
Additionally, we provide the time in seconds required to train each model and the final number of knots used for each.

\subsection{Boston Housing Data} \label{s:boston}

%\jarad{If you decide you want to use Ames data, there is an R package called AmesHousing alternatively there are links to two csv files in the .tex file}
% https://github.com/jarad/jarad.github.com/blob/master/courses/stat587Ag/slides/Ch12a-1-Story%20Houses%201946%20%26%20Newer%20(Detached).csv}
% https://github.com/jarad/jarad.github.com/blob/master/courses/stat587Ag/slides/Ch12a-2-Story%20%26%201-1_2-Story%20Houses%201946%20%26%20Newer.csv

The first data set that we consider is the Boston housing data set\footnote{http://lib.stat.cmu.edu/datasets/boston} \citep{harrison1978}. 
As in \cite{garton2020oat}, 
we use ``\% lower status of the population'', 
``average number of rooms per dwelling'' and 
``pupil-teacher ratio by town'' to predict the median value of owner occupied homes. 
We also removed observations where the median value was less than \$50,000, leaving 490 observations. For each of five runs, we randomly selected $\approx 80\%$ of the data for training and used the remaining $20\%$ for prediction.
%\jarad{Why not do this in the standard cross-validation way where you split the data up into 5 chunks and leave one chunk out at a time?}
%\nate{No reason. By the time I realized that I wasn't doing this in the normal CV way, I already had most of the results, and I 
% couldn't convince myself that there was anything very wrong with doing it this way.}
In addition to the four models mentioned in Section \ref{s:experiments}, 
this data set is small enough that we can easily fit the full GP. 
Additionally, to more accurately provide results for what is currently common practice, 
we also provide results for a VFE model
where knots and covariance parameters are found by simultaneous optimization and knots are initialized with k-means clustering.
Table \ref{t:models_boston} provides a summary of the models that we fit for this data set.

\begin{table}[htbp!]
\begin{minipage}{0.85\textwidth}
\centering
\caption{List of models fit to the Boston housing data. The first model in the table is a full GP. }
%\begin{minipage}{0.85\textwidth}
%\jarad{I think you should create informative acronyms instead of models 1-6,
%perhaps FGP, OBVk, ORVk, OBFk, SVk, and SVO. These should then be used in the figure.}
%\end{minipage}
\begin{tabular}{|l|l|l|l|}
\hline
Model & Knot Selection & Approximation & Knot Init.\\
\hline
FGP & - & - & -\\
\hline
OBVk & OAT-BO & VFE & k-means\\
\hline
ORVk & OAT-RS & VFE & k-means\\
\hline
OBFk & OAT-BO & FIC & k-means\\
\hline
SVk & Simult. & VFE & k-means\\
\hline
SVO & Simult. & VFE & OAT-BO\\
\hline
\end{tabular}
\label{t:models_boston}
\end{minipage}
\end{table}

In addition to MNLP and SRMSE, we also measure the difference between predictions resulting from
the full GP and those resulting from the sparse models. For this, we use the average 
univariate Kullback-Leibler divergence (AUKL) (or its log value) between the predictive 
density from the full GP and that of each sparse model. We calculate this as 

\[ 
AUKL = \frac{1}{N_{test}} \sum_{i = 1}^{N_{test}}\int p_{full}(f(\tilde{\bm{x}}_i)|\hat{\theta}, y) \log \frac{p_{full}(f(\tilde{\bm{x}}_i)|\hat{\theta}, y)}{p_{sparse}(f(\tilde{\bm{x}}_i)|\bm{x}^{\dagger}, \hat{\theta}, y)} df(\tilde{\bm{x}}_i).
\]

\begin{figure}[htbp!]
\centering
 \includegraphics[width = 0.8\textwidth, height = 0.6\textheight]{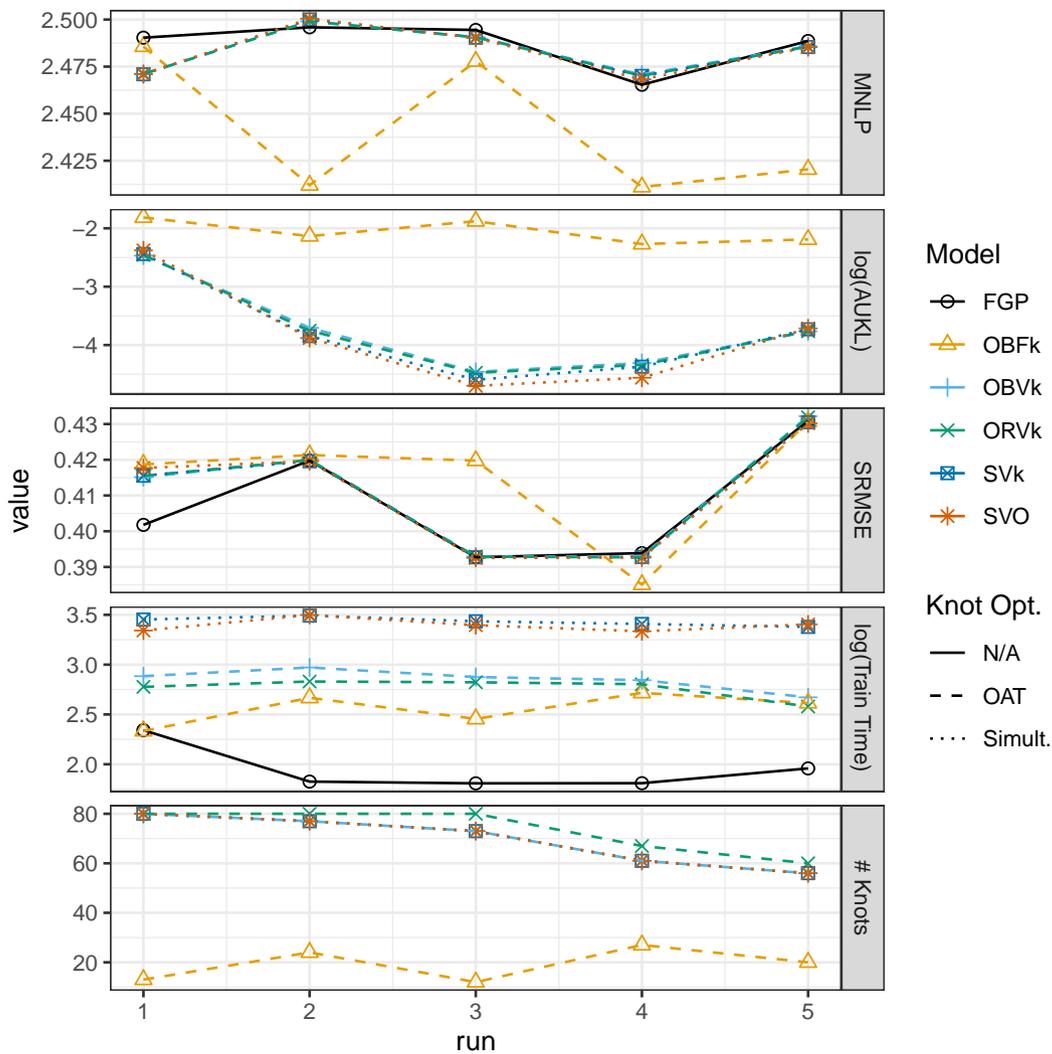}
 \caption{Results on the Boston housing data set for five randomly sampled training and test sets. 
Model enumeration corresponds to Table \ref{t:models_boston}.
%\jarad{Can you change the last facet to be ``\# Knots'' rather than just K?}
 }
 \label{fig:boston_results}
\end{figure}

Figure \ref{fig:boston_results} shows results from each model on each random test set of the Boston data.
Broadly speaking, we see close agreement across all five runs of the accuracy measures for the VFE and the 
full GP models. However, we see that the simultaneously optimized VFE models tend to take two or three times 
longer to fit. Any differences between using the BO and the RS proposal seem to be minimal. 
The FIC model had the largest differences between the other models. For one, it tends to choose models 
with fewer than half as many knots as the VFE models. 
As one might expect, this corresponds to
substantially different predictive distributions compared to the full GP as measured by the (log base 10) AUKL. 
However, it is unclear if the FIC model makes less accurate point predictions
since, other than on the third run, 
the SRMSE values are competitive with each of the other models. 
Furthermore, the FIC MNLP values are smallest for all but the first run 
where MNLP is similar to the other models.
%\jarad{I rewrote some of this paragraph, make sure it still says what you intended it to say.}

\subsection{Airfoil Data} \label{s:airfoil}
In the second experiment, 
we use the Airfoil self-noise data set\footnote{https://archive.ics.uci.edu/ml/datasets/Airfoil+Self-Noise},
which is available from the UCI machine learning repository \citep{uci}.
%\jarad{Generally journals dislike footnotes, we should check to see what our targetted journal says.}
The goal is to predict a component of the overall noise, measured in decibels, 
generated by the airfoil blade of certain aircraft from five continuous 
predictors \citep{gonzalez2008}. 
We fit the same set of models as in the Boston experiment, which are listed 
in Table \ref{t:models_boston}.

%\begin{table}[htbp!]
%\centering
%\caption{List of models fit to the Airfoil self-noise data set.
%\jarad{Since these are the same models as for the Boston housing data, we
% only need one table.}
%}
%\begin{tabular}{|r|l|l|l|l|}
%\hline
%Model & Knot Selection & Approximation & Knot Init. & Abbreviation\\
%\hline
%1 & - & - & - & Full\\
%\hline
%2 & OAT-BO & VFE & k-means & VFE-BO\\
%\hline
%3 & OAT-RS & VFE & k-means & VFE-RS\\
%\hline
%4 & OAT-BO & FIC & k-means & FIC-BO\\
%\hline
%5 & Simult. & VFE & k-means & VFE-Sim\\
%\hline
%6 & Simult. & VFE & OAT-BO & VFE-Ref\\
%\hline
%\end{tabular}
%\label{t:models_airfoil}
%\end{table}

\begin{figure}[htbp!]
\centering
 \includegraphics[width = 0.8\textwidth, height = 0.6\textheight]{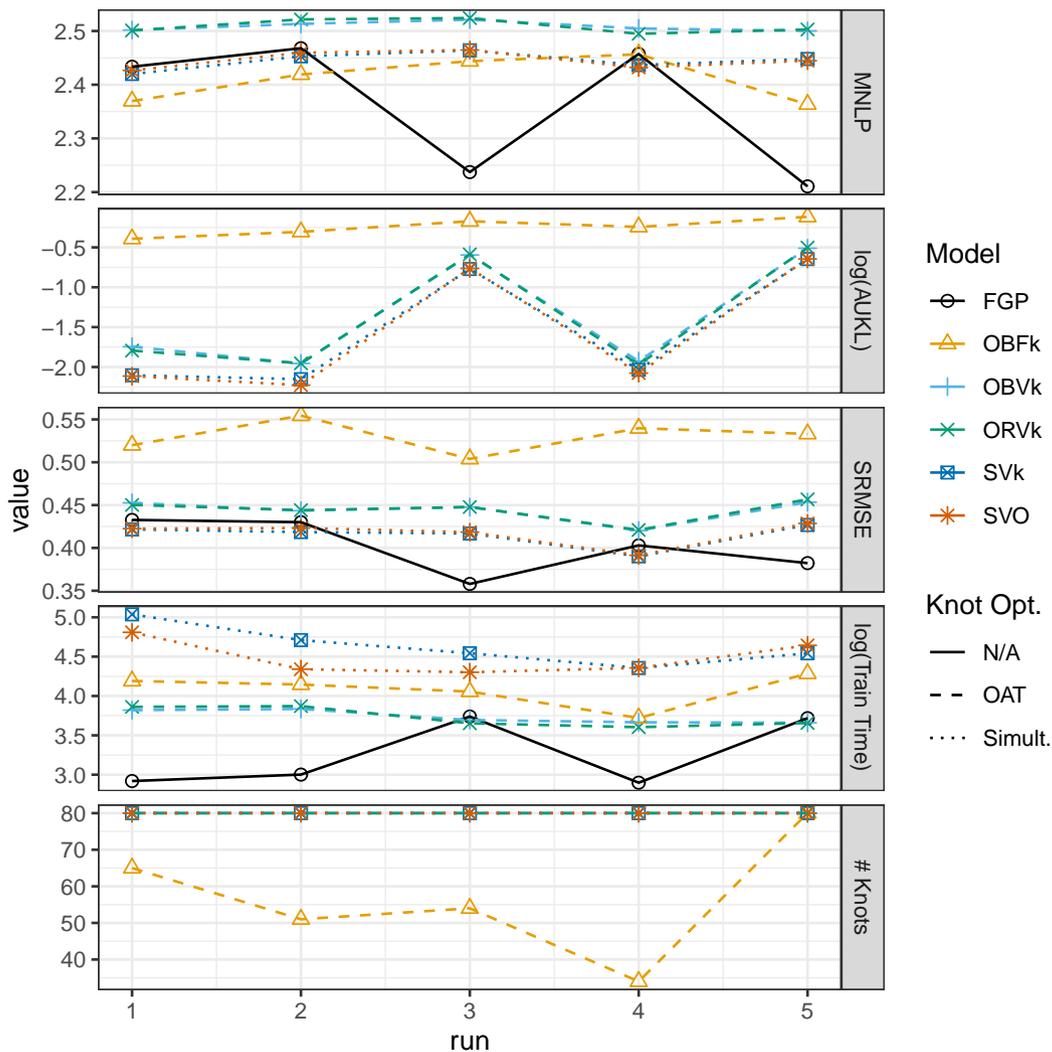}
 \caption{Results on the Airfoil data set for five randomly sampled training and test sets. 
Model enumeration corresponds to Table \ref{t:models_boston}.
%\jarad{Log training time? We should also make sure to comment on training vs testing time. 
%Also, consider putting all figures into the same plot using faceting.}
% \nate{I don't know what there is to say about training vs testing time nor why it is necessary. }
 }
 \label{fig:airfoil_results}
\end{figure}

Figure \ref{fig:airfoil_results} shows results from each model on each random test set of the Airfoil data.
Here, results differ slightly from those on the Boston housing data. We see consistent results for the VFE models chosen via OAT-BO and OAT-RS methods,
but simultaneous optimization seems to result in relatively small, but consistent improvements over the OAT methods.
This improvement comes at an additional computational cost, which is occasionally reduced
through initializing knots to those in the VFE model chosen by the OAT-BO algorithm. The average time
to fit the VFE model with the OAT-BO algorithm was close to 10\% of the average time required by the simultaneously 
optimized VFE model initialized with k-means. Interestingly, while we see the FIC model is again competitive with respect to
the MNLP metric, it now performs consistently worse in terms of SRMSE, explaining roughly $0.5^2 - 0.45^2 = 5\%$ to 
$0.55^2 - 0.45^2 = 10\%$ less variability in the target variable than the VFE models selected using the OAT algorithm. 

\subsection{Combined Cycle Power Plant Data} \label{s:ccpp}
For our third and final experiment, we consider the Combined Cycle
Power Plant (CCPP) data set\footnote{https://archive.ics.uci.edu/ml/datasets/Combined+Cycle+Power+Plant},
which is available from the UCI machine learning repository \citep{uci}. 
The goal is to predict the full load power output of a combined cycle power plant \citep{kaya2012, tufekci2014}. 
The data set consists of 9568 observations of the target variable, power output, along with four other predictor variables.
We randomly split the data five times $\approx$ 50/50 into training and testing sets and provide results for a subset of
the models considered in the previous experiments. 
We do not fit the full GP nor do we fit VFE models with simultaneous
knot optimization where the knot initialization was done with k-means due to time constraints. 
%\jarad{What time constraints? Be more specific or just run them anyway.}
%\nate{This is the sentence in a published paper that we are citing on why they did run an experiment with a full GP
%"Note that the abalone dataset is small enough so as we will be able to train the full GP model". The implication for 
% why they did not run the full GP on other data sets being that they were "too large". There are also experiments that they
% did not run because they were "unrealistically expensive".}
As such, we do not compute the AUKL measure here. 
Table \ref{t:models_ccpp} summarizes the four different models 
fit on each experimental run. 
Model enumeration is kept consistent with the previous experiments for clarity.

\begin{table}[htbp!]
\centering
\caption{List of models fit to the CCPP data set.}
\begin{tabular}{|l|l|l|l|}
\hline
Model & Knot Selection & Approximation & Knot Init.\\
\hline
OBVk & OAT-BO & VFE & k-means\\
\hline
ORVk & OAT-RS & VFE & k-means\\
\hline
OBFk & OAT-BO & FIC & k-means\\
\hline
SVO & Simult. & VFE & OAT-BO\\
\hline
\end{tabular}
\label{t:models_ccpp}
\end{table}

\begin{figure}[htbp!]
\centering
 \includegraphics[width = 0.8\textwidth, height = 0.5\textheight]{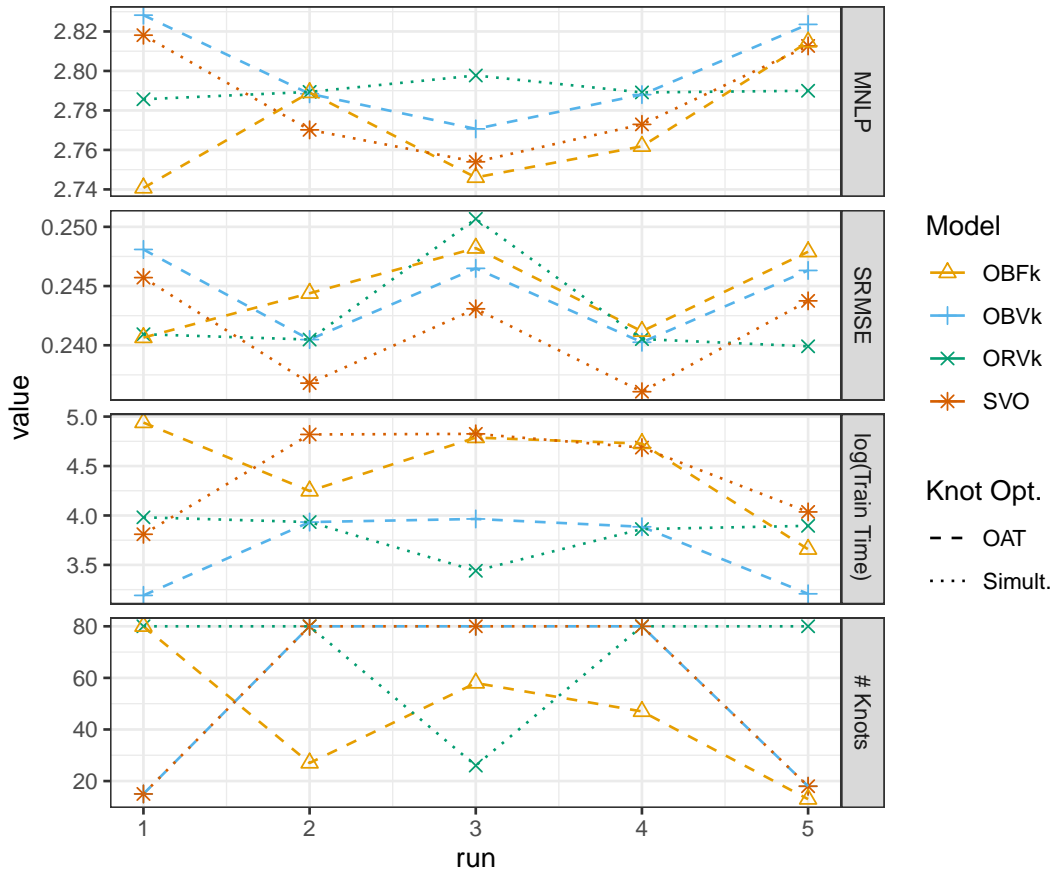}
 \caption{Results on the CCPP data set for five randomly sampled training and test sets. 
Model enumeration corresponds to Table \ref{t:models_ccpp}.
 }
 \label{fig:ccpp_results}
\end{figure}

Figure \ref{fig:ccpp_results} shows results of the four models for the five experimental runs. 
Overall, the four models are similarly accurate with different models achieving MNLP values between roughly 2.74 and 2.83 and
SRMSE values between roughly 0.23 and 0.25 across all five runs.
 Consistent with results on the Airfoil data, we see that simultaneous optimization of 
the knots found by the OAT-BO algorithm in the VFE model results in consistent improvements to the MNLP and
SRMSE values. When the OAT-BO algorithm selects the full 80 possible knots, training time is approximately six to seven times
slower when doing the simultaneous optimization in the VFE model. Surprisingly, despite the FIC model often
having a smaller number knots than the VFE models, training times tended to be roughly comparable to
the simultaneous optimization in the VFE model.

\section{Discussion} \label{s:discussion}
We've tested the OAT knot selection algorithm proposed in 
\cite{garton2020oat} to choose the number 
and locations of knots in the approximate GP regression model proposed by \cite{titsias2009}. 
We compared results on three benchmark regression tasks, 
and found that using the OAT algorithm is always several times faster and 
results in predictions that are competitive with simultaneous optimization of knots. 

\cite{garton2020oat} discussed why the OAT algorithm is typically %CHANGE
faster than simultaneous optimization when the objective function is the marginal 
likelihood, but the same rationale applies here, namely, that gradient evaluations 
cost $\mathcal{O}(dNK^3)$ floating point operations for simultaneous 
optimization and only $\mathcal{O}(dNK^2)$ for the OAT algorithm. 
This difference is even more noticeable as $d$ grows and especially for 
data sets with large $N$. The OAT algorithm does incur additional costs 
due to the knot proposal function, and OAT usually requires a greater 
absolute number of gradient ascent steps. However, these costs are 
usually relatively small in practice. 

Further, \cite{garton2020oat} commented that the simultaneous optimization %CHANGE
of knots with the marginal likelihood as the objective function could result in undesirable 
solutions where several knots serve practically no function. This behavior was also 
discussed in \cite{bauer2016}. OAT has consistently been able to circumvent this problem, and 
this offers a partial explanation as to why OAT may provide competitive or better accuracy when using 
the marginal likelihood as the objective. However, it is noteable that this issue seems far less
 prevalent when the ELBO is used as the objective function. Therefore, why OAT seems to be 
competitive with simultaneous optimization of knots when variational inference is used is less clear. 
With that being said, we make a couple of remarks. First, OAT can be viewed as a kind of 
forward selection algorithm of basis functions in a Bayesian linear nonparametric regression,
 and so the extent to which forward selection algorithms are successful for finding predictive 
linear regression models is likely to be similar here. Second, there are likely many good 
 configurations of knots resulting in very similar predictive distributions.
 We observe this in Figure \ref{fig:gaussian_example}, where none of the knot 
configurations were the same between each model, but model fits were nearly indistinguishable. 
Thus, it seems that significant sophistication may be unnecessary in knot selection algorithms.

We did see that it is sometimes possible to slightly improve the models found using
 the OAT algorithm by refining the knot locations through simultaneous optimization. 
Thus, time permitting, 
one could consider using the OAT algorithm as a way to get a good initialization. 
Further, while we initialized covariance parameters identically in all models
for the sake of comparability, 
we suspect that it would be much faster to initialize covariance parameters to those found
by OAT in the case that OAT is used as an initialization step. 

Interestingly, we did not see substantial differences between using the RS proposal mechanism
and the BO proposal mechanism. 
This is consistent with what was found in \cite{garton2020oat}
when the marginal likelihood was used as the objective function. 
We do find some evidence that when 
a model with few knots can perform well as in, for example, the Boston housing example, using the BO 
proposal tended to select slightly sparser models than the RS proposal. 
This may also have been true of the 
CCPP data, as there the average number of knots selected by the OAT-BO proposal was smaller than the 
average number of knots selected by the OAT-RS proposal, 
but this was not consistent across runs. 
The VFE models using the BO proposal had, on average, 
four fewer final knots than using the RS proposal. 
This makes sense, as the Bayesian optimization should more efficiently 
search candidate knots and avoid local maxima. 
However, in the Airfoil data, where 80 knots were always selected in the OAT models, 
accuracy was indistinguishable between the RS and the BO proposals. 
\cite{garton2020oat} suggested some reasons as to why this BO proposal may not
outperform the RS proposal such as the possibility that the Bayesian optimization spends too much 
time exploring local maxima or that finding a global maximum for a new knot tends to result in a 
final set of knots that is too sparse or clearly suboptimal.

Finally, we also showed how the VFE models compared to the FIC models where optimization was
done through the OAT-BO algorithm. 
When the objective function is the log-marginal likelihood, the OAT
algorithm tends to reliably avoid placing knots directly on top of each other as has been discussed by, for example,
\cite{bauer2016}. 
The OAT-BO algorithm often chooses sparser FIC models than VFE. 
Interestingly, this did not consistently result in either faster training time or reduced accuracy by the measures we considered. 
We do, however, see that the FIC model does not approximate the full GP posterior nearly 
as well as the VFE model does, 
as measured by the KL divergence between the predictive distributions coming from the full GP and
the sparse models. 
The fact that this occurs, but that MNLP and SRMSE values can be competitive with the full GP
and the VFE models suggests that the FIC approximation has utility beyond its ability to mimic a full GP.

With that being said, if the goal of the modeler is to efficiently estimate predictive densities resembling a full GP, then, like
\cite{bauer2016}, our recommendation is to use the VFE approximation over the FIC model. The reason for this 
is that training time in the VFE models is usually at least as short as it is for FIC models, but the VFE 
models appear to more reliably obtain (S)RMSE and MNLP values competitive with a full GP. Furthermore, 
even when FIC models result in good accuracy on the test set, the predictive densities tend to differ from 
the full GP more than the VFE models. 

% It would be interesting to examine the performance of the OAT knot selection routine on
%  variational approximations for classification such as in \cite{hensman2015}, and we hope to pursue 
% this in the future.

\section*{Acknowledgements}
This work was partially funded by the 452 Center for Statistics and Applications
in Forensic Evidence (CSAFE) 453 through Cooperative Agreement \#70NANB15H176 
between NIST 454 and Iowa State University, which includes activities carried out
at 455 Carnegie Mellon University, University of California Irvine, and 456 University of Virginia.  

This work was also partially funded by the Iowa State University Presidential Interdisciplinary Research Initiative on C-CHANGE: Science for a Changing Agriculture.

%\subsection*{Author contributions}

%This is an author contribution text. This is an author contribution text. This is an author contribution text. This is an author contribution text. This is an author contribution text. 

\subsection*{Financial disclosure}

None reported.

\subsection*{Conflict of interest}

The authors declare no potential conflict of interests.

\section*{Supporting information}

The following supporting information is available as part of the online article:

\appendix

\section{Optimal variational distribution derivation}
\label{sec:optimal}
Here we reproduce essentially the same derivation of the optimal variational distribution 
%\jarad{What is the criterion that this is optimal for? 
%It seems like we should be maximizing the ELBO, but, in the appendix, it looks
%like this just provides a bound for the ELBO.}
% \nate{It does maximize the ELBO. The derivation shows that there is an upper bound on the ELBO
% that the ELBO can achieve by setting it to a particular Gaussian.}
and the corresponding ELBO from \cite{titsias2009b}. Note that by ``optimal variational distribution", 
we mean that for the class of approximate posteriors that we consider and for a fixed set of knots, 
we can find the exact approximate posterior that maximizes the ELBO.
Our minor modification to the derivation in \cite{titsias2009b} allows one to arrive at the same approximation in a slightly simpler way. 
We may simply modify our target posterior distribution to be
$p_{GP}(f_{\tilde{\bm{x}}}, f_{\bm{x}}, f_{\bm{x}^\dagger}|y)$ and use a modified class of distributions, $\mathcal{R}$, 
with densities $r$ that can be written as 

\[r(f_{\tilde{\bm{x}}}, f_{\bm{x}}, f_{\bm{x}^\dagger}) = p_{GP}(f_{\tilde{\bm{x}}} , f_{\bm{x}}|f_{\bm{x}^\dagger})h(f_{\bm{x}^\dagger}). \]

\noindent We can then write down the ELBO as follows 

\begin{align*}
ELBO(r) &= E_{r} \left[ \log p(y|f_{\bm{x}})p_{GP}(f_{\tilde{\bm{x}}}, f_{\bm{x}}|f_{\bm{x}^\dagger})p_{GP}(f_{\bm{x}^\dagger}) \right] - E_{r}\left[ \log p_{GP}(f_{\tilde{\bm{x}}} , f_{\bm{x}}|f_{\bm{x}^\dagger})h(f_{\bm{x}^\dagger}) \right] \\
&= E_{r} \left[ \log \frac{p(y|f_{\bm{x}})p_{GP}(f_{\tilde{\bm{x}}}, f_{\bm{x}}|f_{\bm{x}^\dagger})p_{GP}(f_{\bm{x}^\dagger})}{ p_{GP}(f_{\tilde{\bm{x}}} , f_{\bm{x}}|f_{\bm{x}^\dagger})h(f_{\bm{x}^\dagger})} \right] \\
&= E_{r} \left[ \log \frac{p(y|f_{\bm{x}})p_{GP}(f_{\bm{x}^\dagger})}{ h(f_{\bm{x}^\dagger})} \right] \\
&= \int{p_{GP}(f_{\tilde{\bm{x}}} , f_{\bm{x}}|f_{\bm{x}^\dagger})h(f_{\bm{x}^\dagger}) \log \frac{p(y|f_{\bm{x}})p_{GP}(f_{\bm{x}^\dagger})}{ h(f_{\bm{x}^\dagger})} df_{\bm{x}} df_{\tilde{\bm{x}}} df_{\bm{x}^\dagger} } \\
&= \int{p_{GP}(f_{\bm{x}}|f_{\bm{x}^\dagger})h(f_{\bm{x}^\dagger}) \log \frac{p(y|f_{\bm{x}})p_{GP}(f_{\bm{x}^\dagger})}{ h(f_{\bm{x}^\dagger})} df_{\bm{x}} df_{\bm{x}^\dagger} } \\
% &= \int p_{GP}(f_{\bm{x}}|f_{\bm{x}^\dagger}) 
% \left[ \log \frac{p_{GP}(f_{\bm{x}^\dagger})}{ h(f_{\bm{x}^\dagger})} + \log p(y|f_{\bm{x}})\right] 
% h(f_{\bm{x}^\dagger}) df_{\bm{x}} df_{\bm{x}^\dagger}  \\
% &= \int p_{GP}(f_{\bm{x}}|f_{\bm{x}^\dagger}) 
% \left[ \log \frac{p_{GP}(f_{\bm{x}^\dagger})}{ h(f_{\bm{x}^\dagger})} + \log p(y|f_{\bm{x}})\right] 
% h(f_{\bm{x}^\dagger}) df_{\bm{x}} df_{\bm{x}^\dagger}  \\
% &= E_{h} \left[  \int{p_{GP}(f_{\bm{x}}|f_{\bm{x}^\dagger}) \log \frac{p_{GP}(f_{\bm{x}^\dagger})}{ h(f_{\bm{x}^\dagger})} df_{\bm{x}}} + 
% \int{p_{GP}(f_{\bm{x}}|f_{\bm{x}^\dagger}) \log p(y|f_{\bm{x}}) df_{\bm{x}} } \right] \\
&= E_{h} \left[ \log \frac{p_{GP}(f_{\bm{x}^\dagger})}{h(f_{\bm{x}^\dagger})} + \int{p_{GP}(f_{\bm{x}}|f_{\bm{x}^\dagger}) \log p(y|f_{\bm{x}}) df_{\bm{x}} }  \right]. 
\end{align*}

%\jarad{I commented out some intermediate steps that I needed to make sure I understood how you got here. We can consider uncommenting if we think they are helpful.}

\noindent This is the same ELBO as derived by \cite{titsias2009}, and so the same arguments apply to derive the optimal distribution $h^*$. 
The remaining work is replicated from \cite{titsias2009b} with some minor notational differences.
First, we evaluate $\int p_{GP}(f_{\bm{x}}|f_{\bm{x}^\dagger}) \log p(y|f_{\bm{x}}) df_{\bm{x}}$ analytically as follows,
\[ \begin{array}{ll}
\multicolumn{2}{l}{\int p_{GP}(f_{\bm{x}}|f_{\bm{x}^\dagger}) \log p(y|f_{\bm{x}}) df_{\bm{x}}} \\
&= E_{p} \left[ \left. -\frac{N}{2} \log (2\pi \tau^2) - \frac{1}{2 \tau^2} \sum_{i = 1}^{N}{(y_i - f(x_i))^2} \right|f_{\bm{x}^\dagger} \right] \\
&= -\frac{N}{2} \log (2\pi \tau^2) - \frac{1}{2 \tau^2} E_{p} \left[ \left. \sum_{i = 1}^{N}{\Big( \left[y_i - \underbar{m}(x_i)\right] - \left[f(x_i) - \underbar{m}(x_i)\right] \Big)^2} \right| f_{\bm{x}^\dagger} \right] \\
&= -\frac{N}{2} \log (2\pi \tau^2) - \frac{1}{2 \tau^2} \left[ \sum_{i = 1}^{N}{ \left(y_i - \underbar{m}(x_i)\right)^2} + Tr\left( \Sigma_{\bm{x} \bm{x}} - \Sigma_{ \bm{x} \bm{x}^\dagger} \Sigma_{\bm{x}^\dagger \bm{x}^\dagger}^{-1} \Sigma_{\bm{x}^\dagger \bm{x}} \right) \right] \\
&\equiv \log G(f_{\bm{x}^\dagger}, y),
\end{array} \] 

\noindent where $\underbar{m}(x_i) \equiv E_{p}\left[ f(x_i) | f_{\bm{x}^\dagger} \right]$, and expectations are with respect to $p_{GP}(f_{\bm{x}} | f_{\bm{x}^\dagger})$.
In the future, it will be useful to note that $\log G(f_{\bm{x}^\dagger}, y) = \log \mathcal{N}(y; \underbar{m}(\bm{x}), \tau^2 I) - \frac{1}{2\tau^2} Tr\left( V\left[ f_{\bm{x}}|f_{\bm{x}^\dagger} \right] \right)$. 

We then note that 

\begin{equation*}
ELBO(r) = \int{h(f_{\bm{x}^\dagger}) \log \frac{G(f_{\bm{x}^\dagger}, y)p_{GP}(f_{\bm{x}^\dagger})}{h(f_{\bm{x}^\dagger})} df_{\bm{x}^\dagger}}.
\end{equation*}

\noindent We now look for a distribution $h$ that achieves an upper bound on the ELBO. We can do this, as explained by \cite{titsias2009b}, by using Jensen's inequality to see that 

\begin{align*}
ELBO(r) &= \int{h(f_{\bm{x}^\dagger}) \log \frac{G(f_{\bm{x}^\dagger}, y)p_{GP}(f_{\bm{x}^\dagger})}{h(f_{\bm{x}^\dagger})} df_{\bm{x}^\dagger}} \\
&\leq \log \int{ G(f_{\bm{x}^\dagger}, y)p_{GP}(f_{\bm{x}^\dagger}) df_{\bm{x}^\dagger}} \\
&= \log \left[ \mathcal{N}(y \-\ ; \-\ m_{\bm{x}} , \Psi_{\bm{x} \bm{x}} + \tau^2I )  - \frac{1}{2\tau^2} Tr\left( V\left[ f_{\bm{x}}|f_{\bm{x}^\dagger} \right] \right)  \right], 
\end{align*}

%\jarad{I think the inequality is the wrong direction which means this is a lower bound,
%as stated in Titsias (2009b).}
% \nate{You might be thinking about Jensen's inequality for convex functions. Since log is concave, 
% the inequality is in the other direction. If we can find a distribution, h^*, that achieves the preceeding upper 
% bound on the ELBO, then that distribution is optimal.}

\noindent where, recall that we've defined $\Psi_{\bm{x} \bm{x}} =  \Sigma_{\bm{x}^\dagger \bx}  \Sigma_{\bm{x}^\dagger \bm{x}^\dagger}^{-1}  \Sigma_{\bx \bm{x}^\dagger}$. Jensen's inequality becomes an equality when $\frac{G(f_{\bm{x}^\dagger}, y)p_{GP}(f_{\bm{x}^\dagger})}{h(f_{\bm{x}^\dagger})}$ is a constant,
and this occurs when $h(f_{\bm{x}^\dagger}) \propto \mathcal{N}(y; \underbar{m}(\bm{x}), \tau^2 I) p(f_{\bm{x}^\dagger})$. The term on the right hand side of the proportionality sign
can be viewed as a joint distribution for $(Y, f_{\bm{x}^\dagger})$ resulting from a Gaussian likelihood with a Gaussian
prior on the mean. Further, note that the analytically tractable posterior for $f_{\bm{x}^\dagger}$ given $y$ in this model is proportional to the joint distribution, 
and thus works as a choice for $h(f_{\bm{x}^\dagger})$. Thus, we set 

\[ h^*(f_{\bm{x}^\dagger}) = \mathcal{N}(m_{\bm{x}^\dagger} + \Sigma_{\bm{x}^\dagger \bx} \left[ \Psi_{\bm{x} \bm{x}} + \tau^2 I \right]^{-1}(y - m_{\bm{x}}), \Sigma_{\bm{x}^\dagger \bm{x}^\dagger} - \Sigma_{\bm{x}^\dagger \bx} \left[ \Psi_{\bm{x} \bm{x}} + \tau^2 I \right]^{-1} \Sigma_{\bx \bm{x}^\dagger} ). \]

\noindent Using the fact that this choice for $h$ is, in fact the posterior distribution for the model 

\begin{align*}
Y|f_{\bm{x}^\dagger} &\sim \mathcal{N}(\underbar{m}_{\bm{x}}, \tau^2 I) \\
f_{\bm{x}^\dagger} &\sim \mathcal{N}(m_{\bm{x}^\dagger}, \Sigma_{\bm{x}^\dagger \bm{x}^\dagger}),
\end{align*}

\noindent with marginal likelihood $Y \sim \mathcal{N}(m_{x}, \Psi_{\bm{x} \bm{x}} + \tau^2I)$, it is trivial to show that this choice of
$h$ achieves the upper bound on the ELBO and is therefore optimal. Moreover, we have shown that the ELBO is, in fact, equal to

\[ ELBO(r^*) = \log \left[ \mathcal{N}(y \-\ ; \-\ m_{\bm{x}} , \Psi_{\bm{x} \bm{x}} + \tau^2I )  - \frac{1}{2\tau^2} Tr\left( V\left[ f_{\bm{x}}|f_{\bm{x}^\dagger} \right] \right)  \right], \]

\noindent where we use $r^*$ to denote the optimal variational distribution.

%\nocite{*}% Show all bib entries - both cited and uncited; comment this line to view only cited bib entries;
\bibliography{genref_dissertation}%

%\section*{Author Biography}

%\begin{biography}{\includegraphics[width=60pt,height=70pt,draft]{empty}}{\textbf{Author Name.} This is sample author biography text this is sample author biography text this is sample author biography text this is sample author biography text this is sample author biography text this is sample author biography text this is sample author biography text this is sample author biography text this is sample author biography text this is sample author biography text this is sample author biography text this is sample author biography text this is sample author biography text this is sample author biography text this is sample author biography text this is sample author biography text this is sample author biography text this is sample author biography text this is sample author biography text this is sample author biography text this is sample author biography text.}
%\end{biography}

\end{document}